# Development of a Novel Robot for Transperineal Needle-based Interventions: Focal Therapy, Brachytherapy and Prostate Biopsies.

Jean-Alexandre Long, MD (1,2), Nikolai Hungr, PhD (2), Michael Baumann,PhD (2), Jean-Luc Descotes, MD,PhD (1), Michel Bolla, MD (3), Jean-Yves Giraud, PhD(3), Jean-Jacques Rambeaud, MD(1), Jocelyne Troccaz, PhD(2,1)

1- Urology Department, Grenoble University Hospital
2- TIMC-GMCAO Lab (CNRS UMR 5525)
3- Radiotherapy department, Grenoble University Hospital

Word count: 2726
Abstract: 208 words

**Address for correspondence**:
Dr Jean-Alexandre LONG
Department of Urology
Grenoble University Hospital
38043 Grenoble Cedex 9
FRANCE
JALong@chu-grenoble.fr
Tel: + 33 4 76 76 59 70     fax: +33 4 76 76 59 70

Herein is a described a novel robotic system which is the first system using intraoperative prostate motion tracking and MRI-TRUS fusion to guide needles into the prostate. Its use is dedicated to brachytherapy, focal therapy and transperineal prostate biopsies.

**Acknowledgements**: This work was funded by the French Agence Nationale de la Recherche (ANR) under the "Technologies pour la Santé" program (Prosper project, coordinated by Koelis SAS, La Tronche, France). J.Troccaz is supported by INSERM (Contrat Hospitalier de Recherche Translationnelle 2010).

# ABSTRACT

**Purpose:** To report on the initial experience with a new 3D ultrasound robotic system for prostate brachytherapy assistance, focal therapy and prostate biopsies. Its ability to track prostate motion intra-operatively allows it to manage motions and guide needles to predefined targets.

**Materials and methods**: A robotic system for TRUS-guided needle implantation combined with intraoperative prostate tracking was created. Experiments were conducted on 90 targets embedded in 9 mobile and deformable synthetic prostate phantoms. The experiments involved trying to insert glass beads as close as possible to targets in multimodal anthropomorphic imaging phantoms. The results were measured by segmenting the inserted beads in CT scan volumes of the phantoms.

**Results:** The robot was able to reach the chosen targets in phantoms with a median accuracy of 2.73 mm, with a median prostate motion of 5.46 mm. Accuracy was better in apex than in base (2.28 vs 3.83 mm, $p<0.001$) and was similar for horizontal and angled needle inclinations (2.7 vs 2.82 mm, $p=0.18$).

**Conclusion**: This robot for prostate focal therapy, brachytherapy and targeted prostate biopsies is the first system using intraoperative prostate motion tracking to guide needles into the prostate. The preliminary experiments described show its ability to reach targets in spite of the motion of the prostate.

**INTRODUCTION**

Numerous methods exist for treating localized prostate cancer. Among these, radical prostatectomy, external beam radiotherapy and prostate brachytherapy are considered standard treatments, while focal therapies such as focal cryotherapy and photodynamic therapy, although more recent, have been shown to have the potential to improve functional outcomes after treatment of a localized cancer [1,2]. Except HIFU, most of these treatments use implanted needles. Brachytherapy and ablative treatments use a template to insert the needles along a grid of horizontal holes, the depth of each needle being adjusted visually using 2D trans-rectal ultrasound (TRUS) guidance. Limitations to this implantation technique may exist. Both the insertion of the needles and the movement of the probe cause significant motion and deformation of the prostate [3], which could cause implantation inaccuracies. Needle insertion is also restricted to the template's horizontal axis, thereby restricting access behind the pubic arch and limiting eligibility to patients with prostates < 60 cm3 [4,5].

We propose a computer-assisted robotic system called PROSPER, that consists of a robotic needle insertion device and a static 3D ultrasound probe[6-8] (Figure 1). The robot allows needles to be inserted whatever the implanted angles, prostate size and prostate mobility, hence improving implantation accuracy, allowing for targeting of tumor foci and extending the indications for larger prostates. Although focal therapy is the long term goal for the system, brachytherapy was chosen as the first application because of the current lack of long term data concerning focal therapy[9]. This device could also be available for multiparametric prostate biopsies.

This paper presents the robot characteristics and procedure. We report the results of tests to assess the accuracy of needle insertion in prostate phantoms.

## MATERIALS AND METHODS

**ROBOT DESIGN**

The robotic needle manipulator consists of a needle holder that is mounted on the lateral side of the US probe. It positions the needle along the appropriate insertion axis and drives it to a given depth. The robot uses seven motors, giving it seven degrees of freedom **(figure 2).** To increase safety, a mechanical release system disengages the needle from the robot in case the latter comes in contact with the pubic bone. The needle guide at the front of the needle insertion module and the bushing holding the needle are sterilizable. The rest of the robot is covered by a sterile plastic cap. The robot has been designed and developed by our group and described in detail in a previous paper[6].

**CLINICAL PROCEDURE DESCRIPTION**

An Ultrasonix™ 4DEC9-5 end-fire 3D probe is connected to an Ultrasonix RP™ machine. The robotic needle manipulator is rigidly calibrated pre-operatively to the probe placed in the rectum. A 3D US reference volume is acquired. The needle trajectories and seed positions are then defined with respect to the reference prostate, according to the expected planning.

The robot positions the needle-holder in front of the perineum and inserts the first needle. In case of pubic arch interference, the needle is withdrawn and a partial re-planning is done to modify the needle trajectory in order to avoid the pubic arch. Once the needle has been inserted to its planned position, a verification procedure is applied to check for any prostate motion or deformation caused by the insertion. This is done by taking a second 3D US volume and registering it with the reference volume. If the target is not reached, the needle depth is adjusted iteratively until the expected point is reached. The treatment is applied, and the procedure is repeated for the rest of the needles.

**PHANTOM TESTS**

We performed a set of simulated procedures by depositing glass beads through inserted needles into realistic, mobile synthetic prostate phantoms. The results were measured by segmenting the inserted beads in CT scan volumes of the phantoms. Glass beads were chosen instead of titanium seeds because of the absence of artifacts in the CT images.

We developed the phantoms to obtain realism in US, CT and MRI images, but also in the mechanical soft-tissue behavior during needle insertion. The phantoms were built of soft PVC plastic, and included a hard perineum, soft peri-prostatic region, a harder prostate with an echogenic capsule and a hollow rectum for probe insertion. The differences in elasticity of the various anatomical regions made for realistic needle insertion forces and prostate motions. The prostate was molded with ten 1mm diameter glass beads embedded inside, acting as targets for needle insertion. The US probe was inserted into the phantom's rectum, giving realistic images of the prostate, with the glass targets being easily segmented. The phantom and its targets could also be seen clearly in CT images, allowing us to verify placement in CT volumes **(figure 3).** A detailed description of the phantoms has been published elsewhere [10].

The experiment conducted with the PROSPER system involved trying to insert needles as close as possible to 90 targets embedded in 9 phantoms. As this study is an exploratory trial, no validated information on the effect of size and on the variability of needle accuracy was available, thus the sample size had no statistical rationale. We planned initially to construct 10 phantoms, each with 10 target beads embedded inside. Unfortunately, given the complexity of building an anthropomorphic synthetic prostate phantom, one phantom was damaged and was not available for the study.

The experimental procedure was as follows: the targets were segmented in a reference 3D US image and their coordinates were sent to the robot, which proceeded to insert the respective needle accordingly. An 18 gauge Mick Ripple-Hub™ needle was used, with an insertion speed of 5mm/s and a rotation speed of 8rps. Needle rotation is known to decrease needle-tissue forces, needle bending and tissue deformation [11]. Once the needle was inserted, a second US volume was acquired and the initial reference volume registered to it. The new deformed target location was sent back to the robot. Then, it adjusted the needle depth to the closest point along the needle axis to this new target location. This was repeated until no further depth change could be made. A glass bead was then deposited through the hollow needle to mark the needle's final position.

Targets that were near the superior surface of the prostate were approached at a 10° horizontal and vertical inclination to simulate pubic arch avoidance. The phantoms were then imaged in a Philips™ Brilliance 64 CT scanner at a scanning resolution of 0.15 x 0.15 mm per pixel and 0.33 mm slice spacing. The target and inserted beads were segmented and the distance between them was measured. Data were presented as median and interquartile range. Phantom coordinates were categorized into locations (anterior-posterior, right- left-central zone). Comparison of the accuracy according to the locations was performed using a Mann-Whitney U test or a Cruskal-Walis test as appropriate with PASW v18 (SPSS Inc, Chicago,Il).

# RESULTS

In the phantom experiments, at the apex, the median distance between the centers of the target and inserted beads was 2.28mm, while at the base it was 3.83mm ($p<0.001$). The more the targets were on the left, the higher the error was (2.2 vs 3.6 mm, $p=0.002$). The median amount by which the target moved in the needle insertion direction after image registration was 4.0 mm at the apex and 6.5 mm at the base of the prostate (median 5.46 mm, IQR 3.77-6.71). This can be understood as the amount of prostate motion in the needle-insertion direction. A large majority of the targets required one single depth correction to reach the final insertion point, while about 5% required two or more corrections. Accuracy was not different between horizontal and angled needle inclinations (2.7 vs 2.82 mm, $p=0.28$) and between anterior and posterior locations (2.7 vs 2.7 mm, $p=0.59$). The maximum corrections provided by the tracking system were found in left (7.1 mm), base (6.5 mm) and posterior locations (5.9 mm) **(table 1)**. Homogeneous motions of the prostate were detected by the tracking algorithm in all three spatial directions, x, y and z. This suggested, not just prostate translation in the needle direction (z), but also prostate rotation, that was accentuated with the depth of the needle (punctures at the base provide a bigger motion than those at the apex, whatever the axis considered) **(table 2).**

## DISCUSSION

An important aspect for successful transperineal needle-based interventions is the ability to track prostate motion intra-operatively. Deformations and motions are highly underestimated in conventional practice. Artignan et al. estimated that the prostate moved by at least 5 and 3 mm respectively in a sagittal and coronal plane, related only to the probe insertion [12]. Moreover, according to Roy, needle insertion could result in motions of the base of the prostate of up to 15 mm [13]. It has been stated in the literature that the primary axis of prostate mobility during an insertion is along the needle insertion axis [14]. Our experiments confirmed that prostate rotation can also affect the results: the deeper is the insertion, the more the prostate rotated. It was also noticed that peripheral needles caused more rotation than central needles, as could be expected, if the prostate is assumed to be constrained by the pubo-prostatic ligaments.

Clinical outcomes of this imprecision may be negligible in a treatment of the whole gland like for brachytherapy, but they may be more serious in focal therapy or targeted biopsies.

The results described above show that in our synthetic phantoms, needle insertion caused significant motion, on the order of 5mm in the direction of the needle axis. Without the registration step, the resulting seed distribution would have been significantly offset from the planned distribution. By correcting the needle depth based on the prostate motion, the accuracy of the system in the needle insertion direction was measured to be 2.28 mm at the apex and a little more (3.83 mm) at the base due to poorer image resolution at greater depths. The experiments showed that the system was capable of drastically reducing the errors caused by prostate mobility in the cranial-caudal direction. This was true for the different depths of insertion and approach angles tested.

We defined optimal needle insertion and rotation speed according to a needle insertion force study previously published[6]. In the phantom (similarly to in vivo force measurements reported in the literature during live brachytherapy)[15], the maximal force (1.8N) was reached when traversing the perineum. The amount of prostate motion was minimal with an insertion speed of 5mm/s. Although this speed is lower than during a manual procedure, it allows a safe insertion. Indeed, the insertion can be interrupted at any time by pressing the emergency stop button.

The reason why the robot is less precise when targeting to the left of the prostate is unclear. A possible explanation is that the weight of the robot might create increased gravitational strain on the robot causing a curvature towards the left side and thus a lower accuracy. This will have to be improved in the upcoming version of the robot.

The accuracy results are hardly comparable with systems previously described. Indeed, PROSPER is the only robot providing a motion tracking system and angulated needle insertion. As a result it could improve pubic collision avoidance, providing a better access to anterior and lateral parts of the prostate and extending the indications for larger prostates[16].

Conceptually, focal therapy has the potential to minimize treatment-related toxicity without compromising cancer-specific outcome [2]. A current limitation to the proliferation of focal therapy as a viable treatment option is the inability to stage or grade the cancer accurately because of suboptimal imaging capabilities. In the near future, there is no doubt that imaging will detect better cancer foci, especially those with high probability to develop [17]. As shown in recent studies, multiparametric MRI makes it possible to predict more accurately the location of tumor foci particularly those with high grade [18]. We believe that MR imaging could improve the efficiency of focal therapy and obviously prostate biopsies. MRI guidance is a challenging task due to the need of MR compatible instruments and the difficulty to process images in real time. In contrast, MRI-TRUS fusion has the advantage of combining pre-operative MR images and intra-operative ultrasound images in a real-time display[19, 20].

The robot described herein is an attempt at a multimodality robot that could combine the accuracy of robotic needle guidance, precise tracking of prostate motions and deformations as well as MRI-TRUS fusion for facilitated target detection and intervention planning. Indeed, PROSPER is coupled to a MRI-TRUS fusion algorithm. A 3D US reference volume is acquired then registered to a pre-operative MRI acquisition to facilitate and improve prostate delineation, initial dose planning as well as the identification of suspicious areas. Note that MRI fusion was not used in the experiments reported in this paper, because it could interfere with the evaluation of the prostate tracking, which was our main objective. The fusion technique has, however been detailed in previous publications by our group [7, 21].

As previously mentioned, this system has been described initially for a brachytherapy application. Indeed brachytherapy for low-risk localized prostate cancer is a standard of care, while results for focal therapy require more maturity. With this in mind, a dosimetry planning software taking into account the oblique directions of the needles is in process.

The system could, also, be used for prostate transperineal biopsies, as the described system attempts, in general, to help the surgeon to perform accurate punctures in the operating room with the patient in a lithotomic position. Given that TRUS-MRI fusion is available on our system, it could be useful to perform multiparametric prostate biopsies. However, a transperineal approach would be required to use the robot.

The originality of the PROSPER robot is its ability to track prostate motions and deformations to correct the trajectories and depth of the needles[22]. During a manual procedure, especially in brachytherapy, these motions are suspected and the dosimetry plan is manually registered with the prostate shape. This study shows that these motions are not so easily noticeable and that positioning errors may occur also because of prostate rotation, which cannot be evaluated with classic US modalities.

During a prostate brachytherapy procedure, such a robot might be clinically useful to correct for tissue deflection and deformation, adjust seed positions based on real-time dosimetry feedback and avoid the pubic arch so as to extend indications for large prostates. Furthermore, a reduction of the number of needles required in the target volume could be enabled by inverse optimization to create more conformal treatment plans.

These goals can be achieved by oblique needle insertions that require a robotic system. Indeed, manual insertion requires placing needles into areas according to the standard template holes.

Additionally the tracking system has the ability to make fine positional adjustments to offset distribution depending on gland motion.

This robot, by providing a fusion between MRI images and an accurate 3D-TRUS guidance, could represent a future solution for robotic targeted focal therapy [1, 7, 23]. In the mean time, the system can be useful for performing more accurate robotic brachytherapy and prostate biopsies.

The study has limitations. It can be considered as a first experience with this new robot. In the statistical analysis of our phantom experiments, it is important to keep in mind that the number of phantoms used was chosen empirically.

In our model, we did not consider the potential bleeding due to the needle insertion in vivo that can usually cause imaging artifacts, however, each needle inserted into the phantom left a prominent trace in the image, progressively degrading US image quality. This degradation did not affect the US registration algorithm, so we are confident that bleeding will not significantly affect system accuracy.

An important future step for the furthered success of our prostate needle insertion system would be to determine the degree to which the prostate rotation occurs in vivo and to provide ways of mitigating this error, such as predicting the motion with biomechanical models or reducing prostate rotation with stabilizing needles[24].

Before using this system on human beings, preclinical studies on more realistic models are needed. These experiments encourage us to continue on with an in depth pre-clinical cadaver study.

# CONCLUSION

This robot for prostate focal therapy, brachytherapy and targeted prostate biopsies is the first system to use intraoperative prostate motion tracking to guide needles into the prostate. The preliminary experiments described show its ability to reach targets in spite of the motion of the prostate.

**Figure 1:** Schematic representation: a- the PROSPER robot design, b- System setup

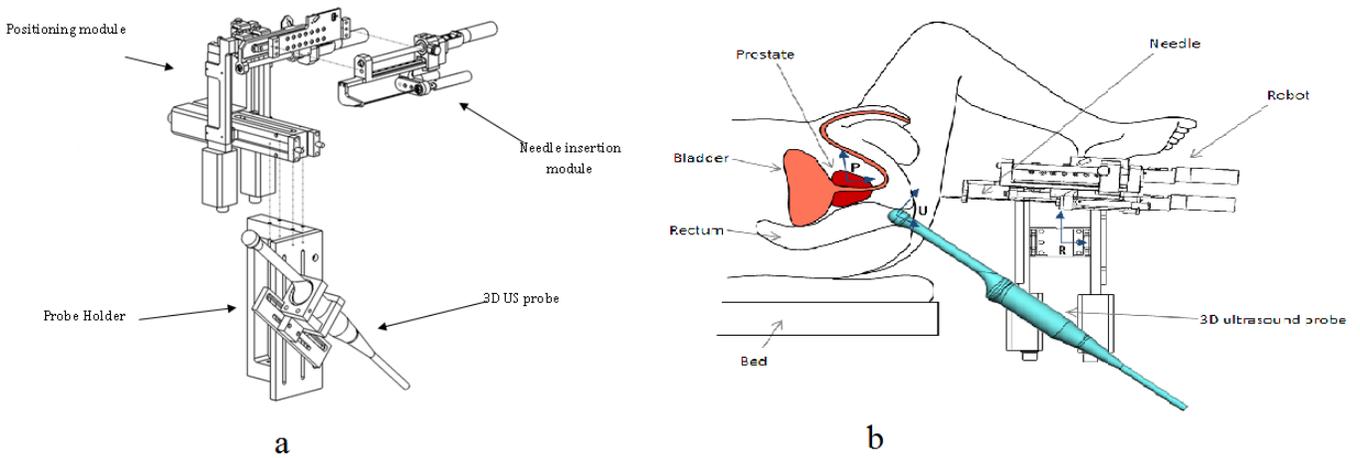

**Figure 2**: (a) Test-bench setup showing all the components of our system (1: 3D endfire US probe, 2: prostate phantom, 3: US machine, 4: needle insertion module, 5: needle positioning module, 6: laboratory robot-probe stand). (b) Photograph of our first robot prototype (7: vertical motors, 8: horizontal motors, 9: Z-translation motor, 10: homing Hall sensors, 11: needle disengagement mechanism, 12: needle insertion motor, 13: needle rotation motor, 14: needle grip, 15: needle, 16: robot end effector).

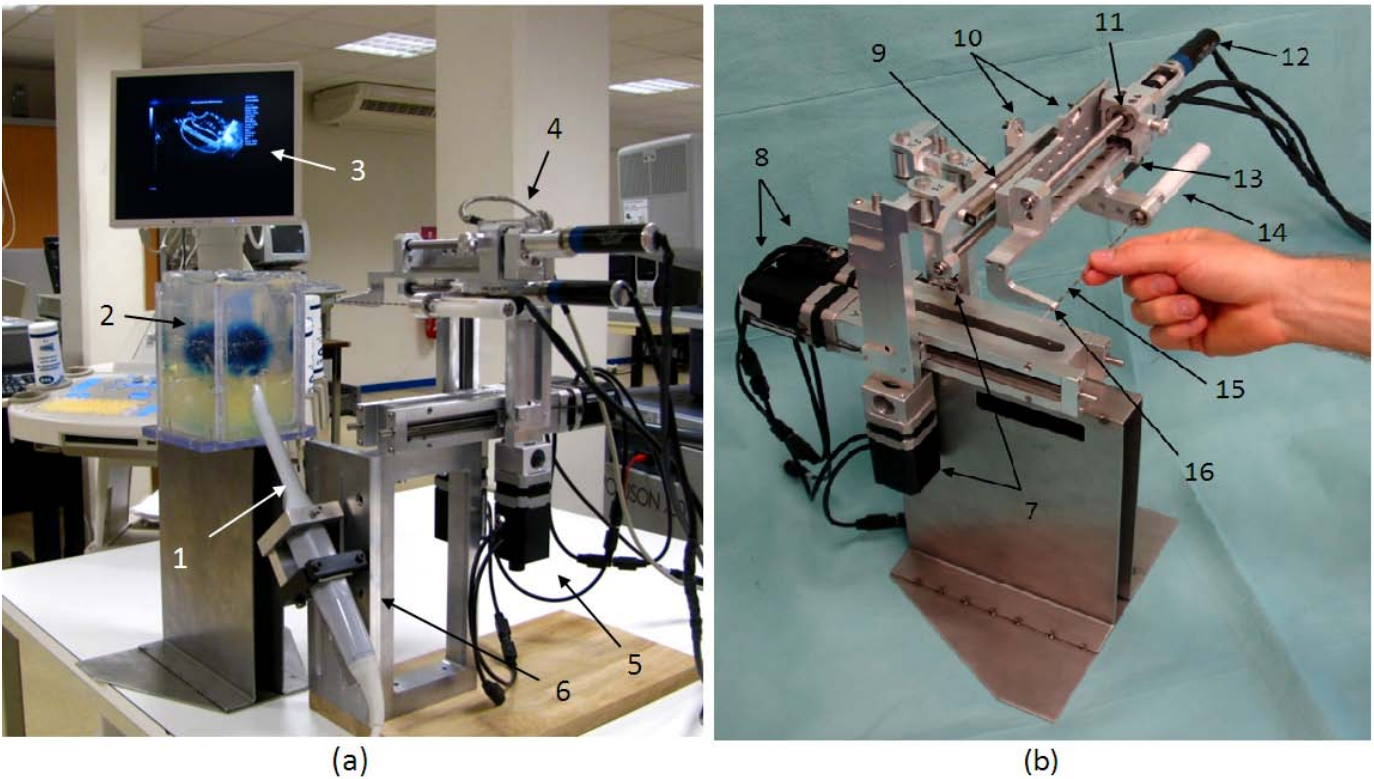

**Figure 3**: Glass bead implantation in a mobile and deformable phantom
  a-US view of the beads embedded inside the phantom with definition of a target
  b-CT scan showing the target hit by a glass bead

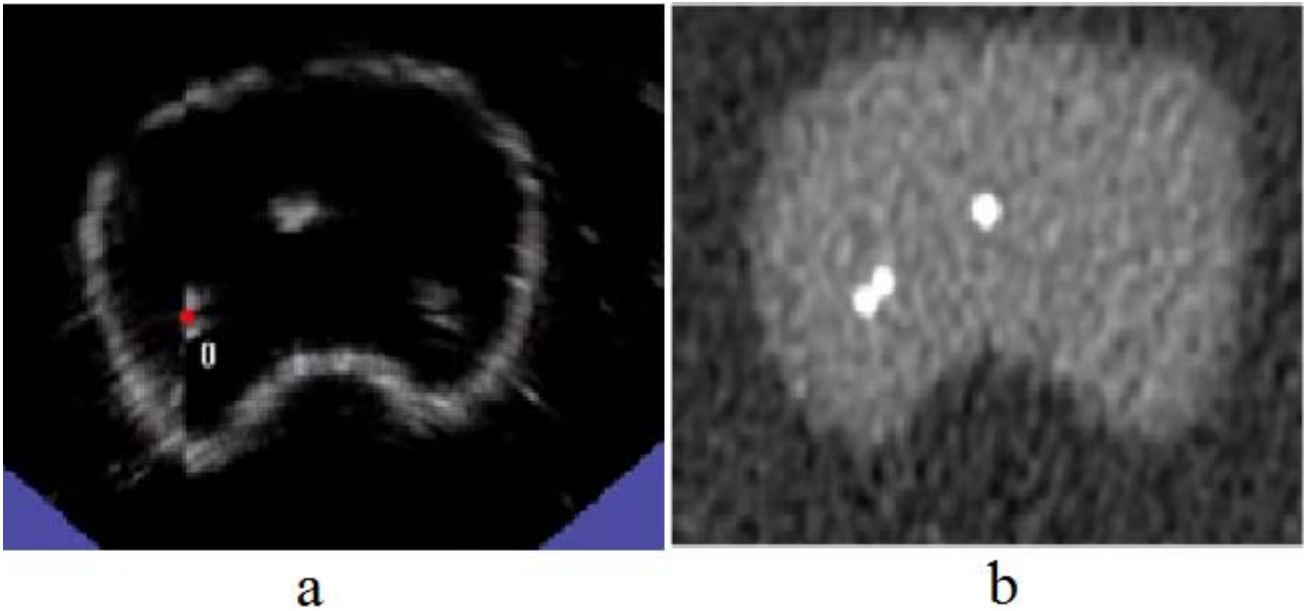

Table 1: Distance errors between targets and beads stratified according to spatial location

|            | # beads | Distance error | p        | Depth correction | p        |
|------------|---------|----------------|----------|------------------|----------|
| **Apex**   | 50      | 2.28 (1.8-2.7) | **<0.001*** | 4.0 (3.0-5.6)    | **<0.001*** |
| **Base**   | 40      | 3.83 (2.9-4.7) |          | 6.5 (5.8-8.2)    |          |
| **Center** | 28      | 2.6 (2.1-3.4)  | **0.002*** | 5.9 (3.1-6.9)    | 0.6      |
| **Left**   | 32      | 3.6 (2.7-4.8)  |          | 7.1 (3.4-11.2)   |          |
| **Right**  | 30      | 2.2 (1.8-2.9)  |          | 5.7 (3.8-9.5)    |          |
| **Anterior**  | 52   | 2.7 (2.1-3.6)  | 0.59     | 5.4 (3.1-6.7)    | 0.18     |
| **Posterior** | 38   | 2.7 (2.1-3.9)  |          | 5.9 (4.6- 6.8)   |          |
| **Horizontal** | 67  | 2.7 (2.1-3.8)  | 0.28     | 5.4 (3.1-6.7)    | 0.18     |
| **Angled**    | 23   | 2.82 (1.8-3.3) |          | 5.9 (4.6- 6.8)   |          |

Values are expressed as median (IQR). Unit is mm.

**Table 2: Median distance of target movements along each axis showing mobility in all three directions and suggesting, not just prostate translation in the needle direction (z), but also prostate rotation.**
Results are expressed as median and IQR (mm)

|  | x | y | z |
|---|---|---|---|
| **overall** | 1.26 (0.57-2.34) | 1.09 (0.51-1.78) | 1.53 (0.89-2.25) |
| **Apex** | 1.10 (0.54-1.81) | 0.73 (0.37-1.47) | 1.36 (0.7-1.89) |
| **Base** | 2.1 (0.81-3.78) | 1.49 (0.7-2.14) | 2.0 (1.05-2.57) |
| **Right** | 0.97 (0.37-1.31) | 0.89 (0.49-1.53) | 1.32 (0.7-2.0) |
| **Center** | 2.03 (1.20-3.7) | 1.35 (0.55-2.17) | 1.59 (1.0-2.6) |
| **Left** | 2.13 (1.23-3.77) | 1.45 (0.57-2.3) | 1.69 (1.03-2.4) |
| **Anterior** | 1.1 (0.55-1.81) | 0.73 (0.37-1.7) | 1.36 (0.7-1.89) |
| **Posterior** | 1.62 (0.61-2.7) | 1.09 (0.62-1.7) | 1.21 (0.69-2.11) |